\documentclass[10pt,twocolumn,letterpaper]{article}

\usepackage{cvpr}              % To produce the CAMERA-READY version
\usepackage{multirow}
\usepackage{colortbl}

\usepackage[accsupp]{axessibility}  % Improves PDF readability for those with disabilities.

\definecolor{cvprblue}{rgb}{0.21,0.49,0.74}
\usepackage[pagebackref,breaklinks,colorlinks,allcolors=cvprblue]{hyperref}

\def\paperID{46243} % *** Enter the Paper ID here
\def\confName{CVPR}
\def\confYear{2026}

\title{SD-FSMIS: Adapting Stable Diffusion for Few-Shot \\ Medical Image Segmentation}

\author{
    Meihua Li$^{1\dagger}$, Yang Zhang$^{1\dagger*}$, Weizhao He$^{1}$, Hu Qu$^{1}$, Yisong Li$^{1}$
    \\
    $^{1}$Computer Vision Institute, College of Computer Science and Software Engineering, Shenzhen University
    \\
    {
    \small 
    limeihua2023@email.szu.edu.cn, yangzhang@szu.edu.cn,
    heweizhao2022@email.szu.edu.cn, \{quhu, liyisong\}2023@email.szu.edu.cn
    }
}
\begin{document}

\maketitle
\def\thefootnote{$^{\dagger}$}\footnotetext[1]{Equal Contribution: Meihua Li and Yang Zhang}
\def\thefootnote{$^{*}$}\footnotetext[2]{Corresponding Author: Yang Zhang}
\begin{abstract}
% The ABSTRACT is to be in fully justified italicized text, at the top of the left-hand column, below the author and affiliation information.
% Use the word ``Abstract'' as the title, in 12-point Times, boldface type, centered relative to the column, initially capitalized.
% The abstract is to be in 10-point, single-spaced type.
% Leave two blank lines after the Abstract, then begin the main text.
% Look at previous \confName abstracts to get a feel for style and length.
Few-Shot Medical Image Segmentation (FSMIS) aims to segment novel object classes in medical images using only minimal annotated examples, addressing the critical challenges of data scarcity and domain shifts prevalent in medical imaging. While Diffusion Models (DM) excel in visual tasks, their potential for FSMIS remains largely unexplored. We propose that the rich visual priors learned by large-scale DMs offer a powerful foundation for a more robust and data-efficient segmentation approach.
In this paper, we introduce SD-FSMIS, a novel framework designed to effectively adapt the powerful pre-trained Stable Diffusion (SD) model for the FSMIS task. Our approach repurposes its conditional generative architecture by introducing two key components: a Support-Query Interaction (SQI) and a Visual-to-Textual Condition Translator (VTCT). Specifically, SQI provides a straightforward yet powerful means of adapting SD to the FSMIS paradigm. The VTCT module translates visual cues from the support set into an implicit textual embedding that guides the diffusion model, enabling precise conditioning of the generation process. 
Extensive experiments demonstrate that SD-FSMIS achieves competitive results compared to state-of-the-art methods in standard settings. Surprisingly, it also demonstrated excellent generalization ability in more challenging cross-domain scenarios. 
These findings highlight the immense potential of adapting large-scale generative models to advance data-efficient and robust medical image segmentation.
\end{abstract}    
\section{Introduction}
\label{sec:intro}

With the rapid advancement of artificial intelligence~\cite{he2016resnet,lin2024SDSeg,zhu2024diffews,oquab2023dinov2,he2024apseg,wang2025disfacerep,wang2025facebench}, automated medical image segmentation has emerged as a cornerstone technology, playing a crucial role in a multitude of clinical applications~\cite{sherer2021metrics,rigaud2021automatic}. It empowers early disease detection and facilitates treatment planning tailored to individual patient characteristics, thereby advancing the frontier of personalized healthcare. However, the remarkable success of these deep learning-based models is critically dependent on vast, meticulously annotated datasets for training. This dependency presents a significant real-world obstacle: acquiring large-scale, high-quality, pixel-level annotations across diverse medical domains is notoriously difficult, prohibitively expensive, and exceptionally time-consuming. Furthermore, the clinical deployment of these models is often plagued by detrimental domain shifts, arising from variations in imaging protocols, scanner types, anatomical presentations, or pathologies unseen during training.

% \begin{figure}[t]
%   \centering
%   \includegraphics[width=1\linewidth]{fig/difference.pdf}
%   \caption{Comparison between our proposed method and previous methods. 
%   % (a)先前的方法从有限的支持集中生成单个或多个类原型，并通过特征匹配对查询图像分割。这些模型没有很好的先验为基础，因而仅从相对单一的数据中学习，这通常很脆弱，难以应对复杂变化的情况。(b)我们采用了一个强大的、预先训练好的基础模型（稳定扩散），而不是构建一个新的网络。SD-FSMIS能够引导其庞大的、可泛化的视觉先验，实现了卓越的鲁棒性和泛化能力，特别是在具有挑战性的跨域场景中。此外，我们选择使用先前工作提出方法，生成伪标签进行训练，移除训练对真实标签的需要。
%    (a) Previous methods build task-specific networks from scratch. They generate class prototypes from the limited support set and perform segmentation via feature matching. Lacking strong priors and trained on constrained data, these models are often brittle and struggle with complex visual variations. 
%    (b) Instead of building a new network, we adapt a powerful, pre-trained foundation model. Our framework steers its vast, generalizable visual priors, achieving superior robustness and generalization, especially in challenging cross-domain scenarios.}
%   \label{fig:difference}
% \end{figure}

\begin{figure}[t]
  \centering
  \begin{subfigure}{1\linewidth}
    \includegraphics[width=1\linewidth]{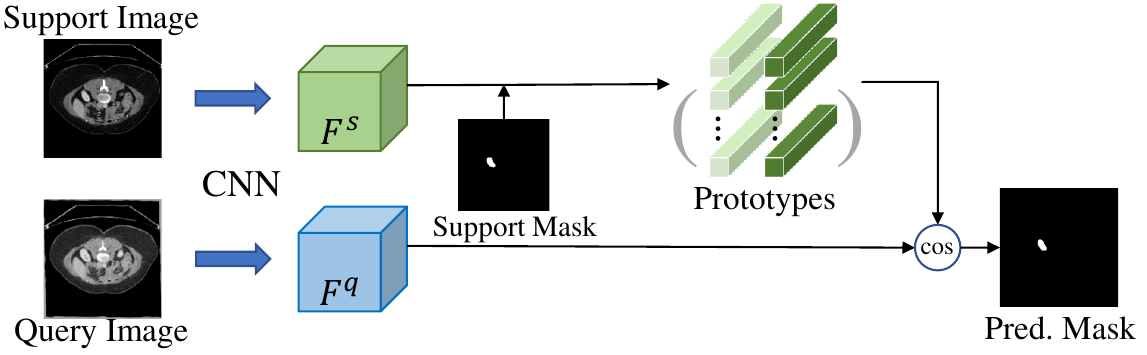}
    \caption{Previous fully supervised methods.}
    \label{fig:difference-a}
  \end{subfigure}
  \hfill
  \begin{subfigure}{1\linewidth}
    \includegraphics[width=1\linewidth]{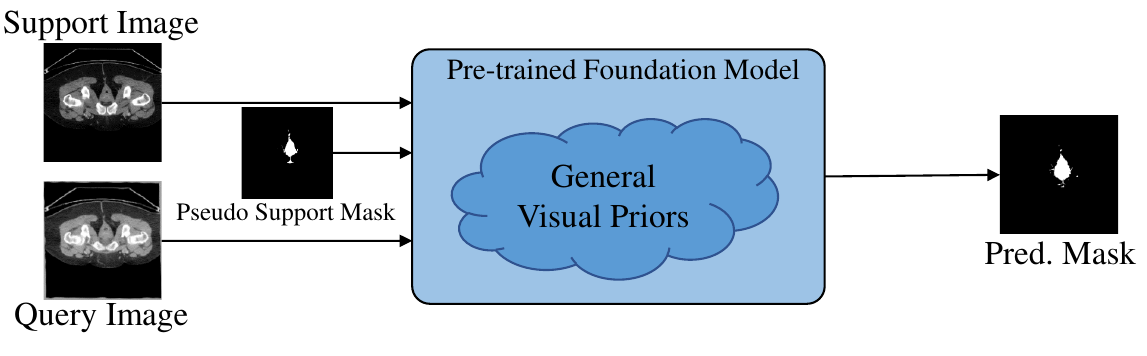}
    \caption{Our method.}
    \label{fig:difference-b}
  \end{subfigure}
  % \caption{Comparison between our proposed method and previous methods. 
  %  (a) Previous methods build task-specific networks from scratch. They generate class prototypes from the limited support set and perform segmentation via feature matching. Lacking strong priors and trained on constrained data, these models are often brittle and struggle with complex visual variations. 
  %  (b) Instead of building a new network, we adapt a powerful, pre-trained foundation model. Our framework steers its vast, generalizable visual priors, achieving superior robustness and generalization, especially in challenging cross-domain scenarios.}
  % 最终版
  \caption{
  Comparison between our proposed method and previous methods. 
  (a) Previous fully supervised methods build task-specific networks from scratch and require pixel-level annotations. They generate class prototypes from the limited support set and perform segmentation via feature matching. Lacking strong priors and trained on constrained data, these models are often brittle and struggle with complex visual variations. 
  (b) Instead of building a new network, we adapt a powerful pre-trained foundation model and do not rely on manual annotations. Our framework steers its vast, generalizable visual priors, achieving superior robustness and generalization, especially in challenging cross-domain scenarios.
  }
  \label{fig:difference}
\end{figure}

Fortunately, Few-Shot Learning (FSL) has emerged as a promising paradigm to address this data scarcity. FSL aims to train a model capable of recognizing and generalizing to novel classes using only a limited number of labeled examples. Consequently, researchers have extended this breakthrough to the medical imaging domain, giving rise to Few-Shot Medical Image Segmentation (FSMIS) ~\cite{zhu2023RPT,cheng2025DIFD} to mitigate the aforementioned challenges. Despite advances driven by meta-learning and prototype-based networks, achieving robust and data-efficient performance remains a significant challenge, especially against the complex and heterogeneous backdrop of medical imaging. The majority of conventional FSMIS approaches still focus on designing more elaborate matching networks, such as those leveraging prototypical networks and attention mechanisms, as illustrated in \cref{fig:difference-a}. Such models, however, are often constrained by their inherent architectural limitations, leading to a performance drop when encountering complex or unseen variations. This brittleness severely limits their clinical utility and robustness.

To bridge this critical generalization gap, we advocate for a paradigm shift, conceptually depicted in \cref{fig:difference-b}. Instead of designing increasingly complex, task-specific architectures trained on limited data, we propose to leverage the powerful and generalizable visual priors encapsulated within large-scale pre-trained foundation models. Diffusion Models (DMs)~\cite{dhariwal2021diffusion}, which have achieved remarkable success not only in generative tasks~\cite{ospa} but have also demonstrated immense potential in fundamental vision tasks like pixel-level prediction~\cite{amit2021segdiff,xu2023openVocabulary} and semantic correspondence~\cite{hedlin2023unsupervised,zhang2023taleTwo}. This potential stems from their ability to learn rich, generalizable representations of texture, shape, and context from massive, diverse datasets such as LAION-5B~\cite{schuhmann2022laion5b}. While these learned priors offer tremendous potential for understanding visual structure, their application to dense prediction tasks like FSMIS remains largely unexplored.

In this paper, we introduce SD-FSMIS, a novel framework that adapts a powerful pre-trained Stable Diffusion (SD)~\cite{rombach2022LDM} model to address the core challenges of FSMIS and enhance its resilience against the complexities of medical imaging. Our central objective is to explore how to efficiently and directly adapt the general-purpose visual priors of SD to serve the FSMIS task. We achieve this through two primary architectural innovations: (1) Support-Query Interaction (SQI), which facilitates effective information exchange in the latent space. By minimally modifying the self-attention layers of the SD model, SQI propagates class-specific information from the support set to the query image, seamlessly adapting SD to the few-shot segmentation paradigm. (2) Visual-to-Textual Condition Translator (VTCT), which acts as a "visual-to-semantic" translator. This module converts class-specific visual information from the support set into implicit text-like embeddings. This allows us to precisely condition the SD model using the "language" it understands, steering its powerful generative priors to focus on the desired anatomical structures while maximally reusing its existing components.

In summary, our main contributions are:
\begin{itemize}
    \item We propose a new paradigm for FSMIS that leverages the rich, generalizable visual priors from pre-trained text-to-image diffusion models to tackle the critical challenge of cross-domain generalization. This shifts the focus from designing task-specific networks from scratch to effectively adapting powerful foundation models.
    \item We introduce SD-FSMIS, a novel yet minimalist adaptation framework. It features the Support-Query Interaction (SQI) module for latent-space fusion and the Visual-to-Textual Condition Translator (VTCT) module, which translates visual cues into text-like conditioning signals to precisely guide the diffusion model.
    \item Extensive experiments demonstrate that our method not only achieves competitive performance in standard FSMIS settings but, more importantly, significantly outperforms state-of-the-art methods in challenging cross-domain scenarios. This empirically validates the superior generalization and robustness of our approach in handling the complex and variable nature of medical imaging.
\end{itemize}

\section{Related Work}
\label{sec:related-work}

\begin{figure*}[!ht]
  \centering
  \includegraphics[width=1\linewidth]{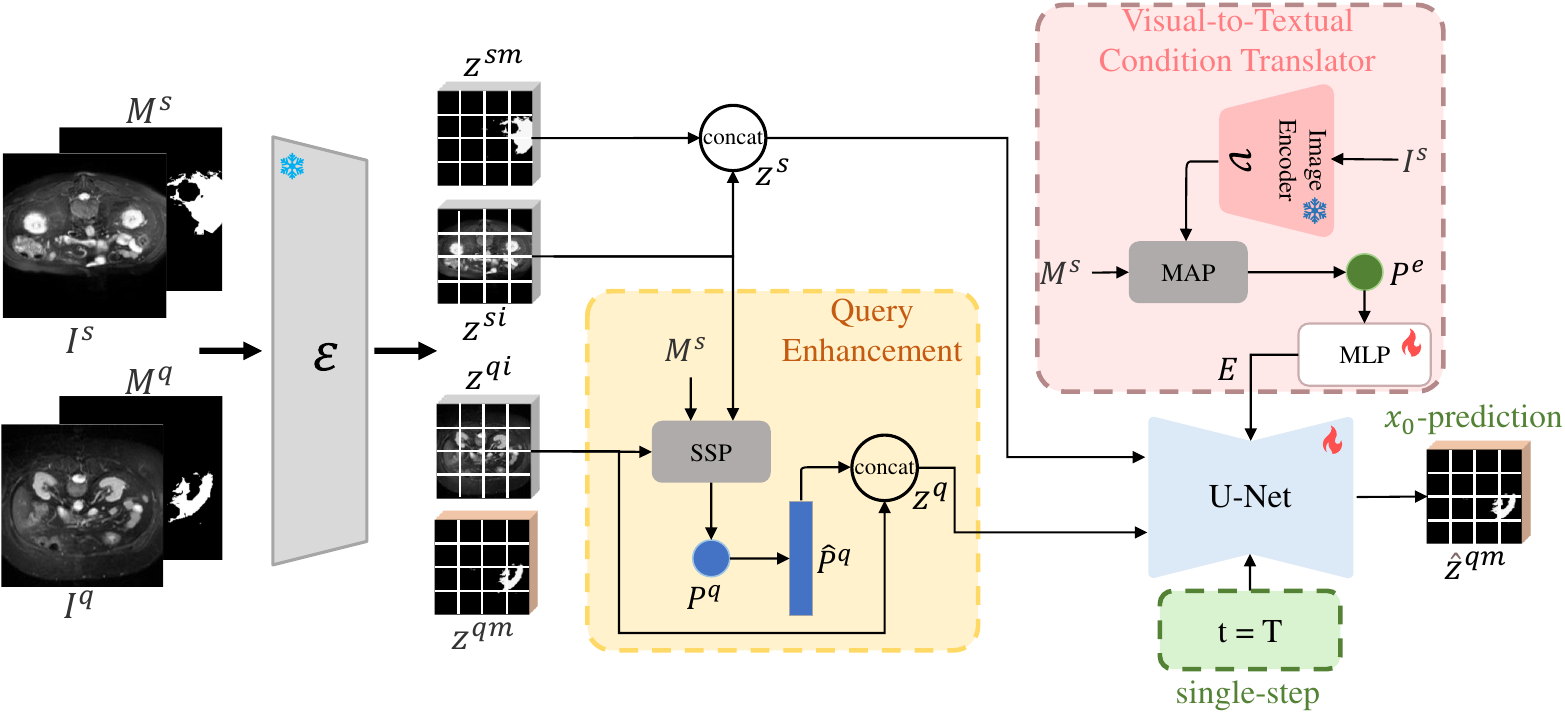}
  \caption{SD-FSMIS Overview and Training Pipeline. Support and query sets are first encoded using the VAE encoder $\mathcal{E}$. The query latent $z^{qi}$ is enhanced via the Query Enhancement module to obtain $z^{q}$, while the support latent $z^{si}$ and its mask latent $z^{sm}$ are concatenated along the channel dimension to form $z^{s}$. These are then fed into the U-Net, where the query mask latent $\hat{z}^{qm}$ is generated under the condition of the text embedding $E$, produced by the Visual-to-Textual Condition Translator module.}
  \label{fig:pipeline}
\end{figure*}

% 小样本语义分割
\textbf{Few-Shot Semantic Segmentation.} 
Few-shot semantic segmentation (FSS) aims to perform dense pixel-level prediction for novel classes guided by a minimal set of annotated support images. The field has been largely shaped by two dominant research thrusts.
The most influential is the prototype-matching paradigm. Pioneered by early works~\cite{shaban2017oneShotSS, dong2018fewShotSS}, this approach generates a representative class prototype from the support set's features, which then guides the segmentation of the query image through a similarity matching process. Subsequent research has focused on refining this core idea; for instance, by creating more discriminative representations through multiple region-aware prototypes~\cite{zhu2023RPT} or by improving the metric learning space~\cite{lang2022BAM, wang2019PANet}. A parallel line of research concentrates on enhancing the interaction between support and query features. These methods aim to achieve a deeper fusion between the two branches, for example, by generating prior maps and leveraging multi-scale feature alignment to enrich the query representation before the final prediction~\cite{tian2020PFENet}.

% Despite their progress, these approaches share a fundamental limitation: they are architecturally designed from the ground up for the FSS task. Consequently, they are trained on relatively constrained datasets and must learn complex visual concepts from scratch. This methodology leads to inherent brittleness, as the learned representations often fail to generalize across significant domain shifts or to visually complex scenes not encountered during training. This exposes a critical gap: the failure to leverage the vast, generalizable knowledge encapsulated in large-scale, pre-trained foundation models. Our work directly addresses this by proposing a new paradigm that adapts, rather than re-designs, to harness these powerful visual priors.

% 小样本医学图像分割
\noindent \textbf{Few-Shot Medical Image Segmentation.} 
The principles of FSS have been adapted to the unique constraints of medical imaging, where pronounced data scarcity and domain heterogeneity are the norms. Existing FSMIS research largely mirrors the trends in FSS, primarily evolving along two axes.
One line of work refines the prototype-matching paradigm to handle the nuances of medical data. Innovations include generating adaptive local prototypes to better capture anatomical variability~\cite{ouyang2020selfFSMIS}, and introducing learnable thresholds to improve the robustness of the matching process against complex backgrounds~\cite{hansen2022ADNet, shen2023QNet}. A parallel effort focuses on designing more elaborate dual-branch interaction mechanisms. These methods employ improved attention or recurrent modules to explicitly model and calibrate the feature-level correlations between support and query images~\cite{roy2020SENet, feng2021MRrNet, ding2023CRAPNet}.
% However, these specialized methods, while innovative, are still custom-built for the medical domain and trained on limited in-domain data. This approach struggles to build truly robust representations, as the models have no exposure to the vast visual world outside of their specific training distribution. The limitations of this paradigm are most starkly revealed in the challenging Cross-Domain FSMIS setting ~\cite{cheng2025DIFD, bo2024FAMNet}. In contrast, our SD-FSMIS utilizes existing visual priors trained on large-scale data with powerful representational capabilities to address the issue of insufficient generalization of models in medical images.
% 相比它们，我们的SD-FSMIS则利用了现有的在大规模数据上训练得到的具有强大表征能力的视觉先验来解决模型对医学图像的泛化性不够的问题。
In this more realistic scenario, the model is tested on target classes and data distributions entirely unseen during training. The difficulty of CD-FSMIS, which is the true litmus test for clinical applicability, underscores the critical need for models endowed with powerful, pre-existing visual priors that are not confined to a narrow medical distribution.

% 扩散模型
\noindent \textbf{Diffusion Models.} 
Diffusion models have demonstrated remarkable capabilities across various visual generation tasks.
Researchers have extensively explored their visual features, applying them to zero-shot classification~\cite{li2023yourDM}, supervised segmentation~\cite{amit2021segdiff}, label-efficient segmentation~\cite{baranchuk2021labelEfficient}, semantic correspondence matching~\cite{hedlin2023unsupervised,zhang2023taleTwo}, and open-vocabulary segmentation~\cite{xu2023openVocabulary}. 
These models typically employ Latent Diffusion Models (LDM)~\cite{rombach2022LDM}, which compress images into latent space, significantly reducing computational costs and enabling the first open-source text-to-image diffusion model at LAION-5B~\cite{schuhmann2022laion5b} scale.
Recent research has shown increasing interest in diffusion-based segmentation methods~\cite{lin2024SDSeg,zhu2024diffews,wu2024medsegdiff,wu2024medsegdiffv2},which generate segmentation predictions based on image conditions. Diffusion models have exhibited substantial potential in fine-grained pixel prediction~\cite{lee2024DMP,xu2023ODISE} and semantic correspondence tasks~\cite{tang2023DIFT,luo2023diffusionHyperfeatures}. 
Notably, SDSeg~\cite{lin2024SDSeg} developed a medical image segmentation approach based on LDM, while DiffewS~\cite{zhu2024diffews} introduced diffusion models to Few-Shot Segmentation, leveraging generative framework priors to maximize task performance.

\section{Methodology}
% 问题表述
\subsection{Problem Formulation}
Few-shot semantic segmentation aims to train a model capable of segmenting novel class images using a minimal number of labeled data, without model retraining. 
Specifically, the training set $D_{train}$ comprises a base class set $C_{train}$ with sufficient annotated samples, and the test set $D_{test}$ includes a novel class set $C_{test}$ with a limited number of annotated samples, where $C_{train} \cap C_{test} = \emptyset$.

We follow the episode-based training approach commonly used in few-shot semantic segmentation tasks ~\cite{wang2019PANet}. The training and testing sets are divided into multiple episodes, each containing a support set $S$ and a corresponding query set $Q$ with the same class. The support set for each class contains $K$ image-mask pairs, denoted as $S = \{(I^i_s, M^i_s)\}^K_{i=1}$, with the corresponding query set represented as $Q = (I^q, M^q)$.
Here, $I \in \mathbb{R}^{H \times W \times 1}$ represents grayscale images, and $M \in \{0, 1\}^{H \times W}$ represents corresponding binary masks. We learn class information from the support set S and then predict masks for query images.

Following ADNet ~\cite{hansen2022ADNet}, we adopt a 1-way 1-shot meta-learning strategy for FSMIS. Additionally, we utilize the supervoxel clustering method proposed by ADNet to generate pseudo-labels as training annotations. This approach allows us to better leverage the volumetric characteristics of medical images and eliminate the need for explicit data labeling.

% 网络结构
\subsection{Network Architecture}
To apply Stable Diffusion to FSMIS, we introduce a novel approach called SD-FSMIS. The overall architecture, illustrated in \cref{fig:pipeline}, comprises two primary novel components: a Support-Query Interaction (SQI) module and a Visual-to-Textual Condition Translator (VTCT) module, which orchestrate the few-shot learning process. The core of our network leverages the powerful generative prior learned by Stable Diffusion, originally trained on the large-scale LAION-5B dataset. We introduce minimal, targeted modifications to adapt its components into an effective few-shot segmentation framework.

\noindent \textbf{VAE Encoder and Decoder.} 
We utilize the pre-trained VAE component of Stable Diffusion to map images and masks into a shared latent space, where the conditional denoising process occurs. The VAE weights are kept frozen throughout our training, preserving the rich visual features learned during its original training. We investigate the reconstruction capabilities of VAE for both medical images and binary masks in the supplementary materials. A key challenge is adapting the VAE, designed for 3-channel RGB inputs, to handle single-channel medical images and their corresponding binary segmentation masks. To address this, we replicate both the input image and the mask across three channels, creating pseudo-RGB representations. Furthermore, pixel values of both image and mask are normalized to the range [-1, 1] to align with the VAE's expected input distribution. During inference, after the diffusion process yields a latent representation of the predicted mask, the frozen VAE decoder maps back to the pixel space. This produces a 3-channel output, which we subsequently average across the channels to obtain the final single-channel predicted segmentation mask.

\noindent \textbf{Adapting U-Net.} 
To adapt U-Net to the input of the support and query latent, we introduced an additional input convolutional layer for cascading features from the support set, following the approach of ~\cite {ke2024Marigold}.

\begin{figure}[!htbp]
  \centering
  \includegraphics[width=0.6\linewidth]{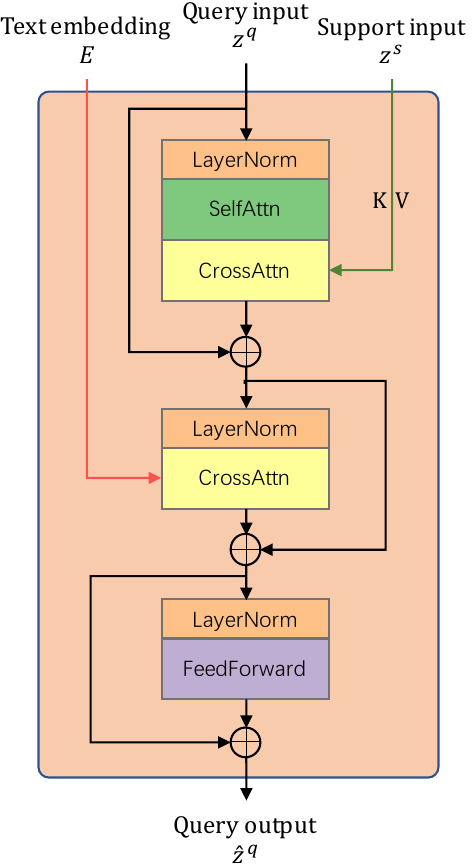}
  \caption{Modified BasicTransformerBlocks architecture. 
  % A CrossAttn module is added after the original SelfAttn, enabling cross-attention between query and support inputs for effective information exchange.
  }
  \label{fig:attn}
\end{figure}

% 支持查询交互
\subsection{Support-Query Interaction}
As illustrated in \cref{fig:pipeline}, the Support-Query Interaction (SQI) module facilitates the integration of support set information into the query feature processing pipeline. We begin by encoding the support set (image $I^s$, mask $M^s$) and the query set (image $I^q$, mask $M^q$) using the frozen VAE encoder $\mathcal{E}$ to obtain their corresponding latent representations $z \in \mathbb{R}^{1 \times c \times h \times w}$. Specifically, we denote these as $z^{si}$ (support image latent), $z^{sm}$ (support mask latent), and $z^{qi}$ (query image latent). The $z^{si}$ and $z^{sm}$ are concatenated channel-wise to form the combined support latent $z^s = concat(z^{si}, z^{sm})$, which serves as input for the U-Net.

\noindent \textbf{Support Information Injection (SII).}
The U-Net in Stable Diffusion utilizes \textit{BasicTransformerBlocks} for integrating conditional information (typically text embedding) with image features. Each block sequentially applies self-attention (SAttn), cross-attention (CAttn), and a feed-forward network (FFN). 
Inspired by the work ~\cite{zhu2024diffews}, we modified this structure to inject support set information to query, as shown in \cref{fig:attn}. After the standard self-attention on the query input $z^{q}$, we introduce an additional cross-attention layer where the query input attend to the support input $z^s$ (acting as key $K$ and value $V$). This enriched query input then undergoes the original cross-attention with the text embedding $E$. The modified operation becomes:
{
\begin{equation}
\hat{z}^{q} = FFN(CAttn(CAttn(SAttn(z^{q}), z^s), E)).
\end{equation}
}

\textbf{Query Enhancement (QE).}
To further enhance the interaction between support and query, we employ a prototype-based query latent enhancement strategy. The architecture of the QE module is highlighted within the yellow block in \cref{fig:pipeline}.

First, we reference the SSP~\cite{fan2022SSP} method to extract the query prototype $P^q$.  Specifically, we compute a foreground prototype $P^s \in \mathbb{R}^{1 \times c}$ from the support set using Masked Average Pooling (MAP) on the support image latent $z^{si}$ guided by the support mask $M^s$:
\begin{equation}
P^s = \frac{\sum_{i,j} M^s_{i,j} \odot z^{si}_{i,j}}{\sum_{i,j} M^s_{i,j}},
\end{equation}
where $M^s_{i,j}$ is the mask value at spatial location $(i, j)$, $z^{si}_{i,j}$ is the corresponding latent feature vector, $\odot$ denotes element-wise multiplication.

Next, we compute the cosine similarity between $P^s$ and $z^{qi}$ to generate a probability map $prob$. We then obtain the query prototype $P^q \in \mathbb{R}^{1 \times c}$ by averaging the query latent $z^{qi}$ for which the corresponding similarity scores $prob$ exceed a threshold $\tau$ (set to 0.7 in our work):
\begin{equation}
P^q = \text{mean} \{ z^{qi}_{i,j} \mid prob_{i,j} > \tau \}.
\end{equation}
This query prototype $P^q$ is expanded spatially to match the dimensions of $z^{qi}$, yielding $\hat{P}^q \in \mathbb{R}^{1 \times c \times h \times w}$. We concatenate this expanded query prototype with the original query image latent along the channel dimension to obtain $z^{qt} \in \mathbb{R}^{1 \times 2c \times h \times w}$: 
\begin{equation}
z^{qt} = concat(z^{qi}, \hat{P}^q).
\end{equation}

% Finally, a $1 \times 1$ convolutional layer followed by a SiLU activation function fuses these features to produce the enhanced query latent $z^q \in \mathbb{R}^{1 \times c \times h \times w}$:
% \begin{equation}
% z^q = SiLU(Conv_{1 \times 1}(z^{qt})).
% \end{equation}

% 隐式文本嵌入生成
\subsection{Visual-to-Textual Condition Translator}
Prior works~\cite{zhu2024diffews} might resort to using null text embeddings, but this provides no specific guidance and fails to leverage the model's powerful text-conditioning mechanism.
To this end, we introduce the Visual-to-Textual Condition Translator (VTCT), a module designed to act as a "visual-to-semantic" bridge. Inspired by ODISE~\cite{xu2023ODISE}, the VTCT's goal is to convert the visual cues from the support set directly into a text-like embedding that the Stable Diffusion model can natively understand.

The architecture of the VTCT module is highlighted within the red block in \cref{fig:pipeline}. To effectively capture the semantic content of the support set, we first employ a pre-trained and frozen image encoder $\mathcal{V}$ to extract features $F^s$ from the support image $I^s$. Subsequently, we perform MAP using the $M^s$ to aggregate the foreground features within $F^s$, yielding a class-specific prototype $P^e \in \mathbb{R}^{1 \times d_{img}}$. Here, $d_{img}$ denotes the feature dimension of the chosen image encoder $\mathcal{V}$.

Finally, this prototype $P^e$, which encapsulates the core visual information of the support class, is fed into a learnable Multi-Layer Perceptron (MLP). This MLP projects $P^e$ into the target embedding space required by the diffusion model's U-Net, producing the implicit text embedding $E \in \mathbb{R}^{1 \times 1 \times d_{text}}$, where $d_{text}$ is the dimension of the text embeddings expected by the U-Net's cross-attention layers. 

This strategy allows us to precisely steer the powerful generative priors of Stable Diffusion towards the desired anatomy by "speaking its language," providing content-aware guidance that is far more specific and effective than a simple null prompt.

% Loss
\subsection{Training Objective}
Our training objective is designed to leverage the strengths of diffusion models for segmentation. In our approach, the model takes an image latent as input and processes it through a U-Net to generate a prediction. The key idea is to train the network to accurately predict the segmentation by comparing the predicted latent with the ground truth mask latent.

Specifically, we use the query mask latent $z^{qm}$, as the target for prediction $\hat{z}^{qm}$. Following the DiffewS~\cite{zhu2024diffews}, we use the Mean Squared Error (MSE) to quantify the difference between the prediction and the target. The loss function is defined as:
\begin{equation}
\mathcal{L} = \frac{1}{h \times w} \sum_{i=1}^{h} \sum_{j=1}^{w}\left(z^{qm}_{i, j} - \hat{z}^{qm}_{i, j}\right)^{2}.
\end{equation}
% This formulation ensures that the network minimizes the difference between its prediction and the negative mask latent, guiding the learning process toward accurate segmentation.

% 推理
\subsection{SD-FSMIS Inference}
\label{subsection:inference}
\begin{figure}[h!]
  \centering
  \includegraphics[width=\linewidth]{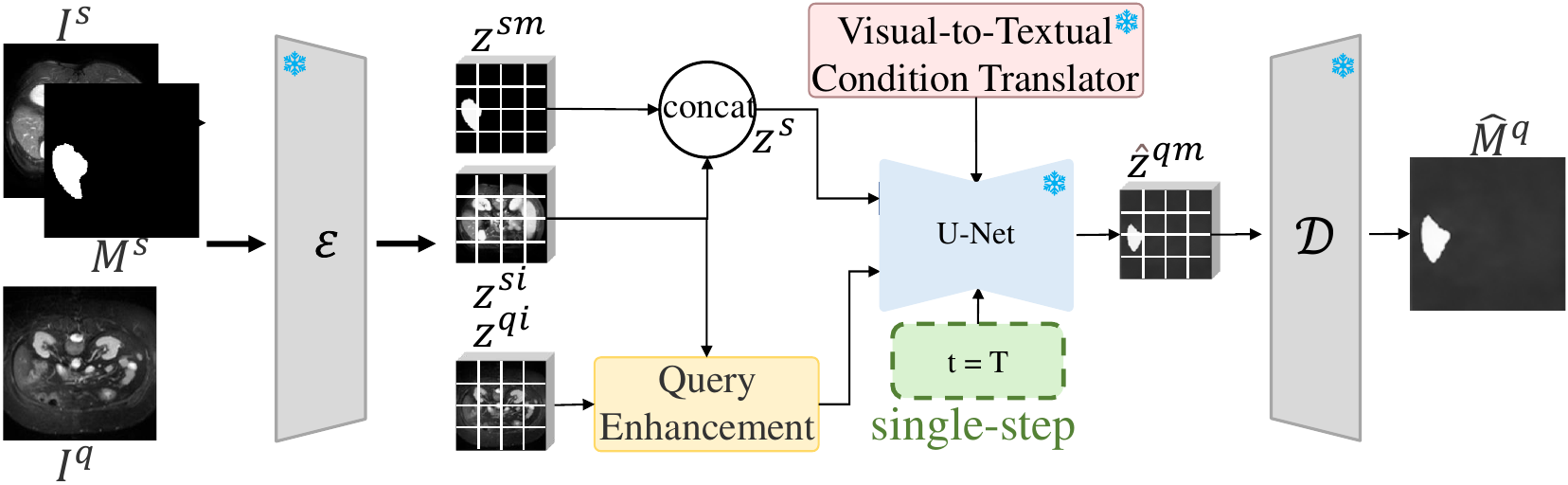}
  \caption{Overview of the SD-FSMIS inference process.}
  \label{fig:inference}
\end{figure}

\cref{fig:inference} illustrates the SD-FSMIS inference process. Specifically, the support and query sets are first encoded into latent space using the VAE encoder $\mathcal{E}$. The support image latent and mask latent are concatenated and fed into the U-Net to provide class information. Under the condition of the generated text embedding $E$, the query latent $z^q$ is segmented in one step and decoded by the decoder $\mathcal{D}$ into an image, with its three channels averaged to produce the final mask $\hat{M}^q$.

\begin{table*}[!ht]
\caption{Quantitative comparison (in Dice score \%) of different methods under setting 1 and 2 on the Abd-MRI and Abd-CT. The best value is shown in bold font, and the second best value is underlined. As "DiffewS" was originally designed for natural images, we re-implemented and trained it on our medical datasets and protocols to ensure a direct and fair comparison.}
\label{tab:result_1}
\centering
\small
\setlength{\tabcolsep}{6pt}{
\resizebox{0.9\linewidth}{!}{
\begin{tabular}{@{}c l l ccccc ccccc@{}}

\toprule

\multirow{2}{*}{Setting} & \multirow{2}{*}{Method} & \multirow{2}{*}{Ref.} & \multicolumn{5}{c}{Abd-MRI} & \multicolumn{5}{c}{Abd-CT} \\
\cmidrule(lr){4-8} \cmidrule(lr{0mm}){9-13} % Partial rules under the main dataset headers
 &  &  & Spleen & Liver & LK & RK & Mean & Spleen & Liver & LK & RK & Mean \\

\midrule

% --- Setting 1 Data ---
\multirow{13}{*}{1} 
 & PANet~\cite{wang2019PANet} & ICCV'19 & 40.58 & 50.40 & 30.99 & 32.19 & 38.53 & 36.04 & 49.55 & 20.67 & 21.19 & 32.86 \\
 & SENet~\cite{roy2020SENet} & MIA'20 & 47.30 & 29.02 & 45.78 & 47.96 & 42.51 & 43.66 & 35.42 & 24.42 & 12.51 & 29.00 \\
 & SSL-ALPNet~\cite{ouyang2020selfFSMIS} & ECCV'20 & 72.18 & 76.10 & 81.92 & 85.18 & 78.84 & 70.96 & 78.29 & 72.36 & 71.81 & 73.35 \\
 & ADNet~\cite{hansen2022ADNet} & MIA'22 & 72.29 & 82.11 & 73.86 & 85.80 & 78.51 & 63.48 & 77.24 & 72.13 & 79.06 & 72.97 \\
 & AAS-DCL~\cite{wu2022AAS-DCL} & ECCV'22 & 76.24 & 72.33 & 80.37 & 86.11 & 78.76 & 72.30 & 75.41 & 74.69 & 74.18 & 73.66 \\
 & Q-Net~\cite{shen2023QNet} & IntelliSys’23 & 75.99 & 81.74 & 78.36 & 87.98 & 81.02 & — & — & — & — & — \\
 & RPT~\cite{zhu2023RPT} & MICCAI’23 & 76.37 & \underline{82.86} & 80.72 & \underline{89.82} & 82.44 & 79.13 & \textbf{82.57} & 77.05 & 72.58 & 77.83 \\
 & CAT-Net~\cite{lin2023CAT-Net} & MICCAI’23 & 68.83 & 78.98 & 74.01 & 78.90 & 75.18 & 67.65 & 75.31 & 63.36 & 60.05 & 66.59 \\
 & PAMI~\cite{zhu2024PAMI} & TMI’24 & 76.37 & 82.59 & 81.83 & 88.73 & 82.83 & 72.38 & 81.32 & 76.52 & \underline{80.57} & 77.69 \\
 & PGRNet~\cite{huang2024PGRNet} & TMI’24 & \textbf{81.72} & \textbf{83.27} & 81.44 & 87.44 & \underline{83.47} & 72.09 & 82.48 & 74.23 & 79.88 & 77.17 \\
 & DIFD~\cite{cheng2025DIFD} & TMI’25 & 75.72 & 80.99 & \textbf{88.59} & \textbf{91.19} & \textbf{84.12} & \underline{79.41} & 79.66 & \underline{83.03} & 78.67 & \underline{80.19} \\

 \cmidrule(lr{0pt}){2-13}

 & DiffewS$^*$~\cite{zhu2024diffews} & NIPS’24 & 76.37 & 78.49 & 75.01 & 87.43 & 79.32 & 78.65 & 80.51 & 74.85 & 80.46 & 78.80 \\ \rowcolor{gray!20}
 & \textbf{Ours} & — & \underline{80.43} & 78.71 & \underline{84.17} & 89.34 & 83.16 & \textbf{85.01} & \underline{81.37} & \textbf{83.21} & \textbf{85.04} & \textbf{83.66} \\
 % & \textbf{Ours} & — & \underline{79.43} & 77.71 & \underline{83.17} & 88.34 & 82.16 & \textbf{83.28} & \underline{81.98} & \textbf{80.02} & \textbf{83.36} & \textbf{82.28} \\
 
\midrule
% --- Setting 2 Data (Values missing in the provided image) ---

\multirow{13}{*}{2} 
 & PANet~\cite{wang2019PANet} & ICCV'19 & 40.58 & 50.40 & 30.99 & 32.19 & 38.53 & 36.04 & 49.55 & 20.67 & 21.19 & 32.86 \\
 & SENet~\cite{roy2020SENet} & MIA'20 & 47.30 & 29.02 & 45.78 & 47.96 & 42.51 & 43.66 & 35.42 & 24.42 & 12.51 & 29.00 \\
 & SSL-ALPNet~\cite{ouyang2020selfFSMIS} & ECCV'20 & 67.02 & 73.05 & 73.63 & 78.39 & 73.02 & 60.25 & 73.65 & 63.34 & 54.82 & 63.02 \\
 & ADNet~\cite{hansen2022ADNet} & MIA'22 & 59.44 & 77.03 & 59.64 & 56.68 & 63.20 & 50.97 & 70.63 & 48.41 & 40.52 & 52.63 \\
 & AAS-DCL~\cite{wu2022AAS-DCL} & ECCV'22 & 74.86 & 69.94 & 76.90 & 83.75 & 76.36 & 66.36 & 71.61 & 64.71 & 69.95 & 68.16 \\
 & Q-Net~\cite{shen2023QNet} & IntelliSys’23 & 65.37 & 78.25 & 64.81 & 65.94 & 68.59 & — & — & — & — & — \\
 & RPT~\cite{zhu2023RPT} & MICCAI’23 & \underline{75.46} & 76.37 & 78.33 & 86.01 & 79.04 & 70.80 & 75.24 & 72.99 & 67.73 & 71.69 \\
 & CAT-Net~\cite{lin2023CAT-Net} & MICCAI’23 & 67.31 & 75.02 & 75.31 & 83.23 & 75.22 & 66.02 & 80.51 & 68.82 & 64.56 & 70.88 \\
 & PAMI~\cite{zhu2024PAMI} & TMI’24 & 75.80 & \textbf{81.09} & 74.51 & 86.73 & 79.53 & 71.95 & 74.13 & 72.36 & 67.54 & 71.49 \\
 & DGPANet~\cite{shen2024DGPANet} & TMI’24 & 69.21 & 75.93 & 73.76 & 75.96 & 73.72 & 65.91 & 65.56 & 74.10 & 68.06 & 68.41 \\
 & DIFD~\cite{cheng2025DIFD} & TMI’25 & 70.96 & \underline{80.38} & \textbf{85.38} & \textbf{90.54} & \underline{}{81.82} & \underline{78.67} & \underline{80.77} & \textbf{84.47} & \underline{75.48} & \underline{79.85} \\

 \cmidrule(lr{0pt}){2-13}

 & DiffewS$^*$~\cite{zhu2024diffews} & NIPS’24 & 73.11 & 77.16 & 77.41 & 83.47 & 77.79 & 76.84 & 79.57 & 69.70 & 73.62 & 74.93 \\ \rowcolor{gray!20}
 % & \textbf{Ours} & — & \textbf{76.25} & 77.58 & \underline{84.03} & \underline{87.27} & \underline{81.28} & \textbf{83.08} & \textbf{82.59} & \underline{82.22} & \textbf{85.10} & \textbf{83.25} \\
 & \textbf{Ours} & — & \textbf{77.25} & 78.58 & \underline{85.03} & \underline{88.27} & \textbf{82.28} & \textbf{83.08} & \textbf{82.59} & \underline{82.22} & \textbf{85.10} & \textbf{83.25} \\

% \midrule

% Fully & Ours & — & 84.09 & 92.80 & 88.36 & 92.59 & 89.47 & 86.00 & 93.27 & 85.50 & 90.30 & 88.77 \\

\bottomrule

\end{tabular}
 }
}
\end{table*}
\section{Experiments}

% 实验设置
\subsection{Experimental Setup}
\textbf{Datasets.} 
Following the evaluation protocol in RPT~\cite{zhu2023RPT}, we assess the performance of our model on Abd-MRI~\cite{kavur2021CHAOS} and Abd-CT~\cite{landman2015SABS}. 

\noindent \textbf{Evaluation Metric and Settings.} 
We primarily use the Dice Similarity Coefficient (DSC) to measure segmentation accuracy, which is the standard metric for this task. 
All experiments are conducted in the 1-shot setting, and performance is reported as the average over a 5-fold cross-validation to ensure statistical robustness.

To evaluate the generalization capabilities of our method, we adopt two challenging cross-domain settings proposed in prior work. Setting 1: Slices containing test classes might be present in the training set, but only as un-annotated background regions. The model is trained on pseudo-masks. Setting 2: The training slices containing the test classes are removed from the dataset. This ensures model has zero exposure to the target anatomies during training, presenting a more realistic and challenging clinical scenario.

\noindent \textbf{Implementation Details.}
Our framework is built upon the Stable Diffusion v1.5 model. Input images are resized to $256 \times 256$, consistent with previous methods. The image encoder within VTCT module is a DINOv2-small~\cite{oquab2023dinov2}.
% We follow ADNet \cite{hansen2022ADNet} to generate pseudo-masks, which serve as the supervisory signal for adapting the diffusion model.
% 最终版
We follow RPT \cite{zhu2023RPT} to generate pseudo-masks, which serve as the supervisory signal for adapting the diffusion model.
The model is trained for 15k iterations per fold on a single NVIDIA A6000 GPU, with a total training time of approximately 6 hours per fold and a memory footprint of roughly 18GB. We employ the AdamW optimizer with a weight decay of 1e-2 and a batch size of 1. The U-Net is trained with a learning rate of 1e-5, while the learnable MLP layers use a higher learning rate of 5e-5. For the diffusion process, we utilize a single-step DDIM scheduler with the timestep t set to 999.

\begin{figure*}[ht]
  \centering
  \includegraphics[width=0.68\linewidth]{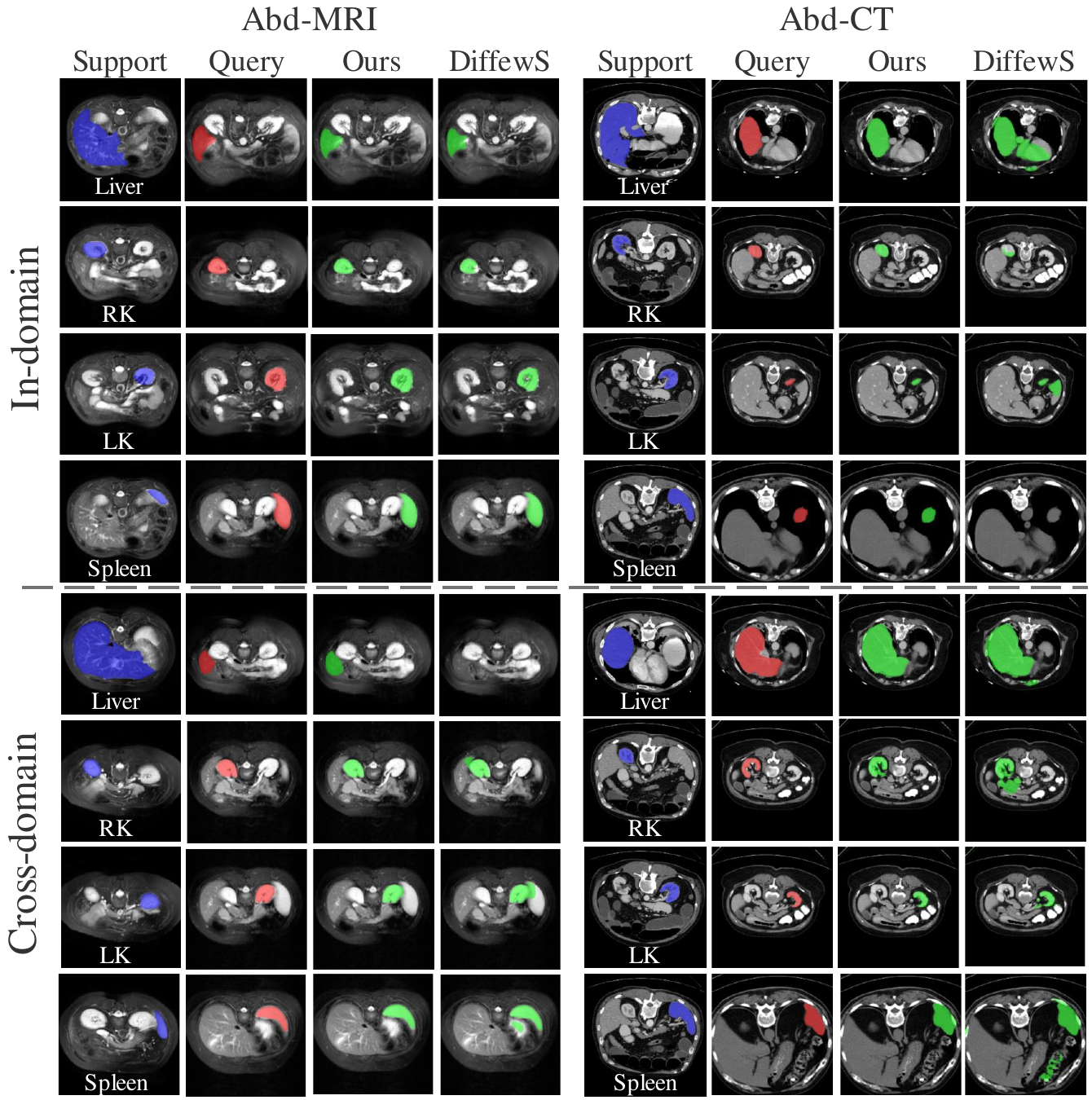}
  \caption{Qualitative comparison between our method and DiffewS method on the Abd-MRI dataset and Abd-CT dataset.}
  \label{fig:vis_result}
\end{figure*}

\begin{table*}[!htbp]
\caption{Quantitative comparison (in Dice score \%) of different cross-domain methods under setting 1. The best value is shown in bold font, and the second best value is underlined. As DiffewS was originally designed for natural
images, we re-implemented and trained it on our medical datasets and protocols.}
\label{tab:result_cross_domain}
\centering
\small
\setlength{\tabcolsep}{6pt}{
\resizebox{0.84\linewidth}{!}{
\begin{tabular}{@{}l l ccccc ccccc@{}} % c for Setting, l for Method, c for data columns
\toprule
\multirow{2}{*}{Method} & \multirow{2}{*}{Ref.} & \multicolumn{5}{c}{Abd-CT $\rightarrow$ MRI} & \multicolumn{5}{c}{Abd-MRI $\rightarrow$ CT} \\
\cmidrule(lr){3-7} \cmidrule(lr{0pt}){8-12} % Partial rules under the main dataset headers
 &  & Spleen & Liver & LK & RK & Mean & Spleen & Liver & LK & RK & Mean \\
\midrule

PANet~\cite{wang2019PANet} & ICCV'19 & 41.83 & 40.21 & 35.36 & 40.32 & 39.43 & 36.58 & 54.63 & 21.95 & 29.19 & 35.58 \\
SSL-ALPNet~\cite{ouyang2020selfFSMIS} & ECCV'20 & 54.01 & 50.62 & 47.30 & 53.07 & 51.25 & 39.23 & 60.80 & 33.01 & 38.24 & 42.82 \\
RPT~\cite{zhu2023RPT} & MICCAI'23 & 54.94 & 52.58 & 42.58 & 58.44 & 52.13 & 52.04 & 59.47 & 40.29 & 49.60 & 50.35 \\
DR-Adapter~\cite{su2024DR-Adapter} & CVPR'24 & 53.91 & 62.79 & 71.67 & 74.12 & 65.62 & 55.77 & 70.83 & 55.93 & 44.20 & 56.68 \\
IFA$_{T=3}$~\cite{nie2024IFA} & CVPR'24 & 59.14 & 67.04 & 73.90 & 75.37 & 68.86 & 56.44 & 71.50 & 54.60 & 50.85 & 58.35 \\
FAMNet~\cite{bo2024FAMNet} & AAAI’25 & 58.21 & 73.01 & 57.28 & 74.68 & 65.79 & 65.78 & 73.57 & 57.79 & 61.89 & 64.75 \\
DIFD~\cite{cheng2025DIFD} & TMI'25 & 61.05 & 68.29 & 75.51 & 79.69 & 71.14 & 62.45 & 73.86 & 56.86 & 46.88 & 60.01 \\

\midrule

DiffewS~\cite{zhu2024diffews} & NIPS’24 & \underline{71.85} & \underline{74.84} & \underline{77.83} & \underline{83.86} & \underline{77.09} & \underline{71.06} & \underline{80.39} & \underline{70.12} & \underline{66.28} & \underline{71.96} \\ \rowcolor{gray!20}
\textbf{Ours} & — & \textbf{74.70} & \textbf{75.26} & \textbf{84.77} & \textbf{90.96} & \textbf{81.42} & \textbf{77.78} & \textbf{81.02} & \textbf{73.03} & \textbf{71.72} & \textbf{75.90} \\
\bottomrule
\end{tabular}
}
}
\end{table*}

% 与其他方法的对比
\subsection{Comparative Analysis with State-of-the-Art}
% 表1展示了我们提出的SD-FSMIS与现有方法的比较，包括基于原型的方法，如Q-Net和DIFD，以及基于交互的方法，例如CRAPNet。为了进行完整的评估，我们还将最近基于扩散的FSS模型Diffews纳入我们的比较中。值得注意的是，由于Diffews最初是为自然图像设计的，我们在医学数据集和协议上重新实施和训练了它，以确保直接和公平的比较。在Abd-MRI数据集上，SD-FSMIS与经过复杂设计的DIFD取得了相当的平均Dice分数。在Abd-CT数据集上，我们的方法明显优于其他方法，在设置1和设置2中分别比SoTA模型高出3.47%和3.4%。DiffewS方法仅对支持和查询使用注意力机制交互，但依靠扩散模型预先学习到的强大视觉表征也能够取得和先前经过精心设计的方法相差不大的效果。我们提出的SD-FSMIS经过适配，比DiffewS的Dice评分平均提高4.88%。我们可以发现SD-FSMIS在设置1和2上的得分相差不大，不会和先前方法出现设置2性能骤降的情况，这说明我们的方法泛化性能更好，能够应对更严苛的医学场景。
% 最近的研究，FAMNet和DIFD都关注到跨域FSMIS（CD-FSMIS）的设置。CD-FSMIS的目标是开发一种通用模型，能够跨显著的领域变化进行推广，例如不同成像方式（如CT与MRI）之间的变化，这严重限制了传统模型的适用性。而在这种具有挑战性的设置中，我们的方法优于先前方法，展示了其出色的泛化能力。性能差距可归因于方法上的根本差异。传统方法学习的表示仅限于其训练数据的有限、相对均匀的分布。当面对一个新的领域时，他们的所学到的先验变得脆弱。相比之下，SD-FSMIS利用了稳定扩散模型中封装的庞大而多样的视觉知识。它对形状、质地和背景等基本概念的理解与特定的医学模式无关。我们提出的适应模块，SQI和VTCT，有效地将这种强大的、预先存在的知识引导到特定的解剖目标上，从而形成一个本质上更稳健、适应性更强的模型。这从经验上验证了将范式从从头开始设计转变为有效适应是解决跨域医学图像分割挑战的更有前景的途径。
\cref{tab:result_1} shows the comparison between our proposed SD-FSMIS and the existing methods. For a complete evaluation, we also include the recent diffusion-based FSS model, DiffewS~\cite{zhu2024diffews}. 
On the Abd-MRI dataset, SD-FSMIS achieves an average Dice score comparable to the intricately designed DIFD~\cite{cheng2025DIFD}. Its advantages become more pronounced on the Abd-CT dataset, surpassing the best method by 3.47\% in Setting 1 and 3.4\% in Setting 2. The DiffewS method employs attention mechanisms solely for the interaction between support and query sets, yet it achieves performance comparable to previously well-designed approaches by leveraging the powerful visual representations pre-learned by the diffusion model. SD-FSMIS demonstrates superior performance after adaptation, achieving an average improvement of 4.88\% in Dice score compared to DiffewS.
Unlike prior methods that often suffer a sharp performance degradation when moving from Setting 1 to the more stringent Setting 2, SD-FSMIS exhibits remarkable resilience. The minimal drop in scores validates its superior generalization and robustness, confirming its suitability for challenging and unpredictable medical scenarios.

% 最近的研究中，FAMNet和DIFD都关注到跨域FSMIS（CD-FSMIS）这种更为复杂的设置，其目标是开发一个通用模型，能够泛化显著的领域变化(例如，从 CT 到 MRI)，这是传统模型难以胜任的任务。在这种具有挑战性的设置中，我们的方法优于先前方法，展示了其出色的泛化能力。我们的结果中显示的性能差距可归因于范式的根本差异。传统方法学习表示仅限于其有限的、相对同质的训练数据，这使得它们在面对新领域时学习的先验知识变得脆弱。相比之下，SD-FSMIS利用了稳定扩散模型中封装的庞大而多样的视觉知识。它对形状、纹理和上下文等基本概念的理解是模态无关的。我们提出的适应模块SQI和VTCT是有效地将这种强大的、预先存在的知识引导到特定解剖目标的关键，从而形成一个本质上更稳健、适应性更强的模型。这从经验上验证了将范式从从头开始设计转变为有效适应是解决跨域医学图像分割挑战的更有前景的途径。
Recent studies have focused on the more challenging cross-domain FSMIS (CD-FSMIS) setting, which aims to develop a universal model capable of generalizing across significant domain shifts (e.g., from CT to MRI). Under this challenging setting, our method demonstrates superior generalization capability, as shown in \cref{tab:result_cross_domain}.
The performance gap shown in our results can be attributed to a fundamental difference in paradigms. Conventional methods learn representations confined to their limited, relatively homogeneous training data, making their learned priors fragile when facing a new domain. In contrast, SD-FSMIS taps into the vast and diverse visual knowledge encapsulated within the Stable Diffusion model. Its understanding of fundamental concepts like shape, texture, and context is modality-agnostic. Our proposed adaptation modules, SQI and VTCT, are the key to effectively steering this powerful, pre-existing knowledge to the specific anatomical target, resulting in a model that is inherently more robust and adaptable. This empirically validates that shifting the paradigm from designing from scratch to effectively adapting is a more promising path for solving the challenges of medical image segmentation.
% 最终版
Additional results compared with universal models~\cite{butoi2023universeg,wong2025multiverseg,wu2024one-prompt} are provided in the supplementary material.
% Additional results with universal models are provided in the supplementary material.

% 消融实验
\subsection{Ablation Study and Visualizations}
% 在这一部分中，我们进行了详细的消融研究，以评估SD-FSMIS框架内每个模块的单独贡献，如表\ref{tab:消融}所示。首先，支持信息注入（SII）模块使扩散模型适应FSMIS任务。仅使用SII，我们的性能与SoTA方法相当，证明了扩散模型的巨大潜力。接下来，我们分别评估查询增强（QE）和隐式文本嵌入生成（ITEG）模块。QE的整合使Dice的平均得分提高了2.16%，而ITEG提高了3.06%。当所有三个组件组合在一起时，该模型的平均Dice得分为83.66%，比SoTA方法提高了3.47%。这些结果证实了SII、QE和ITEG在提高FSMIS任务性能方面的协同作用。
\noindent \textbf{Effect of Each Component.}
We conducted a detailed ablation study to dissect the individual contributions of our framework's key components, as shown in \cref{tab:ablation}.
We first establish a strong baseline by adapting the pre-trained diffusion model using only the SII module. This minimal adaptation, by itself, already achieves performance comparable to existing state-of-the-art methods, demonstrating the immense untapped potential of diffusion model priors for the FSMIS task.
Building on the SII baseline, we integrated the VTCT. This single addition yielded a significant performance increase of 3.06\% in the average Dice score. This confirms that translating visual support cues into text-like conditioning signals is a highly effective strategy for precisely steering the model's generative process.
Similarly, when adding the QE module to the baseline, we observed a 2.16\% improvement. This highlights the importance of facilitating a deeper, more effective fusion of support and query information within the latent space.
Finally, our complete SD-FSMIS framework, which integrates all components, achieves the highest average Dice score of 83.66\%. This represents a 3.47\% gain over the previous state-of-the-art, clearly demonstrating a powerful synergistic effect where each module complements the others to maximize performance.
\begin{table}[!htbp]
\caption{Ablative results of various components of the proposed method on the Abd-CT dataset under setting 1.}
\label{tab:ablation}
\centering
\small
\setlength{\tabcolsep}{5pt}{
\begin{tabular}{cccccccc}
\toprule
SII & QE & VTCT & Spleen & Liver & LK & RK & Mean \\
\midrule
\checkmark &            &            & \textbf{85.54} & 81.86 & 71.72 & 81.31 & 80.11 \\
\checkmark &            & \checkmark & 83.00 & \textbf{82.83} & \underline{81.39} & \textbf{85.45} &  \underline{83.17} \\
\checkmark & \checkmark &            & 83.74 & \underline{81.97} & 80.23 & 83.36 & 82.27 \\
\checkmark & \checkmark & \checkmark & \underline{85.01} & 81.37 & \textbf{83.21} & \underline{85.04} & \textbf{83.66} \\
\bottomrule
\end{tabular}
}
\end{table}

% \begin{table}
%   \caption{Results.   Ours is better.}
%   \label{tab:example}
%   \centering
%   \begin{tabular}{@{}lc@{}}
%     \toprule
%     Method & Frobnability \\
%     \midrule
%     Theirs & Frumpy \\
%     Yours & Frobbly \\
%     Ours & Makes one's heart Frob\\
%     \bottomrule
%   \end{tabular}
% \end{table}

\begin{table}[!htbp]
\caption{Comparison of different versions of Stable Diffusion on the Abd-CT dataset under setting 1.}
\label{tab:sd}
\centering
\small
\setlength{\tabcolsep}{6pt}{
\begin{tabular}{cccccc}
\toprule
Version & Spleen & Liver & LK & RK & Mean \\
\midrule
SD-1.5 & \textbf{85.01} & 81.37 & \textbf{83.21} & 85.04 & \textbf{83.66} \\
SD-2.1 & 83.11 & \textbf{83.02} & 80.01 & \textbf{85.21} & 82.84 \\
\bottomrule
\end{tabular}
}
\end{table}

\noindent \textbf{Version Comparative Analysis.} 
We also evaluated the performance of different versions of the SD model as the backbone for our SD-FSMIS framework. 
\cref{tab:sd} show that SD 1.5 yields superior performance compared to SD 2.1 in our task. We attribute this to their distinct pre-training schemes. The broader, less-filtered dataset used for SD 1.5 appears to provide more generalizable visual priors that, after our adaptation, are better suited for the structural and textural features found in medical scans. In contrast, the heavily filtered dataset and different text encoder of SD 2.1 may result in priors that are less aligned with this specific downstream task. Consequently, we selected SD 1.5 as the default backbone for all our main experiments to ensure optimal performance.

\noindent \textbf{Visualization of results.} 
To further demonstrate the effectiveness of our method, \cref{fig:vis_result} presents visualization results of the predictions from SD-FSMIS on the Abd-MRI and Abd-CT datasets under Setting 1. The results clearly show that our SD-FSMIS generates high-quality segmentation masks for organs with varying intensities, scales, and morphologies, even in complex backgrounds. Furthermore, our model produces high-quality segmentation results even when applied to the more challenging cross-domain FSMIS tasks. 
% This observation reinforces the robustness and generalizability of our approach, demonstrating its ability to handle variations in image distribution and complex segmentation scenarios across different medical imaging modalities.

% \noindent \textbf{Visualization of Training.}
% Additionally, \cref{fig:denoise} illustrates the performance of our method during the training process on the Abd-CT dataset. Notably, even in the early stages of training (after 500 iterations), the model is able to segment simpler classes effectively, and for more complex organs such as the liver, good segmentation results are achieved as early as 5,000 iterations. These findings further underscore the powerful capabilities of diffusion models in tackling few-shot segmentation challenges.

% \input{fig/denoise}

\section{Conclusion}
In this work, we proposed SD-FSMIS, a novel approach that leverages Stable Diffusion for the FSMIS task. To adapt the SD model for FSMIS, we introduced a Support-Query Interaction module that facilitates effective information exchange between support and query. Additionally, we proposed a Visual-to-Textual Condition Translator module to harness the prior knowledge of the SD model by learning text-like embeddings from the support to guide query segmentation. Experiments on Abd-MRI, and Abd-CT datasets demonstrate that SD-FSMIS achieves competitive Dice scores compared to state-of-the-art FSMIS methods. Furthermore, our cross-domain experiments validate the generalization ability and robustness of the proposed approach, yielding the best Dice performance across various scenarios.
\section*{Acknowledgments}
This work was supported in part by the National Natural Science Foundation of China under Grant 62176163, Shenzhen Higher Education Stable Support Program (General Project) under Grant 20231120175215001, and Scientific Foundation for Youth Scholars of Shenzhen University.

\clearpage
\setcounter{page}{1}
\setcounter{section}{0}
\maketitlesupplementary

% ----------- Supplementary Content Starts Here -----------
\section{Implementation Details}
This section provides comprehensive details of our experimental setup to ensure full reproducibility. Our code is implemented based on RPT~\cite{zhu2023RPT} and DiffewS~\cite{zhu2024diffews}.

\textbf{Framework and Model Configuration.} 
Our framework is built upon the publicly available Stable Diffusion v1.5 model. 
All input images are resized to a resolution of 256 $\times$ 256 pixels, maintaining consistency with the protocols of prior FSMIS methods. 
The vision encoder used within our Visual-to-Textual Condition Translator (VTCT) module is a pre-trained DINOv2-small (DINOv2-s) model.

\textbf{Training Data and Supervision.} 
We do not use any ground-truth segmentation masks for training. Instead, we strictly follow the self-supervised strategy proposed in ADNet~\cite{hansen2022ADNet} to generate pseudo-masks, which serve as the sole supervisory signal.
To enhance model robustness and prevent overfitting, we adopt the same data augmentation strategy as RPT~\cite{zhu2023RPT}. Both support and query images undergo random geometric transformations (including rotation, scaling, and translation) and elastic deformations.

\textbf{Evaluation Protocol.}
We evaluate our model following the evaluation protocol implemented in RPT. Specifically, we first select the medical image volume of a single patient from the validation fold of five-fold cross-validation as the support volume (support set), which is then excluded from the validation fold. We leverage this support volume as the support information to perform segmentation on the remaining patients in the validation fold (query set).
During the segmentation process, both the support volume and each query volume are split into three consecutive sub-volumes. The middle slice within each sub-volume of the support volume is utilized to segment all slices in the corresponding sub-volume of the query volume.

\textbf{Training Hyperparameters.} 
% We use the AdamW optimizer with a weight decay of 1e-2. The batch size is set to 1.
% The parameters of the U-Net backbone are fine-tuned with a small learning rate of 1e-5, while our newly initialized and trainable MLP layers use a higher learning rate of 5e-5.
% For the diffusion process, we utilize a single-step DDIM scheduler, following the precise setup in DiffewS~\cite{zhu2024diffews}. The reverse-process timestep t is fixed at 999.
The training process is conducted using the AdamW optimizer with a weight decay value set to 1e-2. The batch size is fixed at 1, and the training is reproducible with a random seed of 42. To stabilize the training, gradient clipping is applied to all parameters with a maximum norm of 1.0.
For the U-Net backbone, we employ a fine-tuning strategy with a relatively low learning rate of 1e-5. In contrast, the trainable MLP layers use a higher learning rate of 5e-5 to facilitate faster convergence.
Regarding the diffusion process, we adopt a one-step DDIM scheduler, following the exact configuration specified in DiffewS~\cite{zhu2024diffews}. Specifically, during the image generation phase, the mask is generated in a single step starting from t=999.

\textbf{Hardware and Training Environment.} 
All experiments were conducted on a single NVIDIA A6000 GPU with 48GB of VRAM.
The model training for each fold consists of 15,000 iterations, which takes approximately 6 hours to complete.
The training process occupies about 18GB of GPU memory.

\section{More Experiments}
\subsection{Validation of VAE Reconstruction Capability}
A fundamental premise of our work is that the Stable Diffusion's pre-trained Variational Autoencoder (VAE) can effectively compress medical images into a meaningful latent space. To validate this, we conducted a direct reconstruction experiment, as the fidelity of reconstruction directly reflects the richness of the encoded visual features.
We passed medical images and their corresponding ground-truth masks through the VAE's encoder and then its decoder. The quality of the reconstructed outputs was quantitatively measured using Mean Squared Error (MSE), Peak Signal-to-Noise Ratio (PSNR), and the Structural Similarity Index (SSIM).
The results, presented in \cref{tab:recon}, show very low MSE and high PSNR/SSIM values for both the images and their masks. This indicates a high-fidelity reconstruction, confirming that the VAE's latent space effectively captures the essential structural and textural features of medical anatomy. This provides a robust and reliable feature foundation upon which our adaptation modules can successfully operate.
\begin{table}[!htbp]
\caption{Quantification of VAE reconstruction quality on Abd-MRI and Abd-CT.}
\label{tab:recon}
\centering
\setlength{\tabcolsep}{6pt}{
\begin{tabular}{@{}cc ccc@{}}
\toprule
Dataset & Type & MSE$\downarrow$ & PSNR$\uparrow$ & SSIM$\uparrow$ \\
% \multirow{2}{*}{Method} & \multirow{2}{*}{Ref.} & \multicolumn{2}{c}{Training} & \multicolumn{2}{c}{Inference} & \multirow{2}{*}{Dice} \\
% \cmidrule(lr){3-4} \cmidrule(lr{0pt}){5-6}
%  &  & Time (hours) & Resources ($\times$ GPUs) & Time (hours) & Speed (sample/s) & \\
\midrule
\multirow{2}{*}{Abd-MRI} & Image & 0.0005 & 34.1592 & 0.9108 \\
\cmidrule(lr{0pt}){2-5}
 & Maks & 0.0007 & 32.2890 & 0.9597 \\
\midrule
\multirow{2}{*}{Abd-CT} & Image & 0.0020 & 27.4889 & 0.8172 \\
\cmidrule(lr{0pt}){2-5}
 & Maks & 0.0009 & 32.0274 & 0.9461 \\
\bottomrule
\end{tabular}
}
\end{table}

\begin{table*}[htb]
\caption{Quantitative comparison (in Dice score \%) of different cross-domain methods under setting 2. The best value is shown in bold font, and the second best value is underlined.}
\label{tab:result_cross_domain_2}
\centering
\small
\setlength{\tabcolsep}{6pt}{
% \resizebox{\linewidth}{!}{
\begin{tabular}{@{}l l ccccc ccccc@{}} % c for Setting, l for Method, c for data columns
\toprule
\multirow{2}{*}{Method} & \multirow{2}{*}{Ref.} & \multicolumn{5}{c}{Abd-CT $\rightarrow$ MRI} & \multicolumn{5}{c}{Abd-MRI $\rightarrow$ CT} \\
\cmidrule(lr){3-7} \cmidrule(lr{0pt}){8-12} % Partial rules under the main dataset headers
 &  & Spleen & Liver & LK & RK & Mean & Spleen & Liver & LK & RK & Mean \\
\midrule

PANet~\cite{wang2019PANet} & ICCV'19 & 33.57 & 31.93 & 27.10 & 32.08 & 31.17 & 28.12 & 41.78 & 16.72 & 20.78 & 26.85 \\
SSL-ALPNet~\cite{ouyang2020selfFSMIS} & ECCV'20 & 51.12 & 47.75 & 44.34 & 50.23 & 48.36 & 34.89 & 54.37 & 30.06 & 33.91 & 38.31 \\
RPT~\cite{zhu2023RPT} & MICCAI'23 & 52.70 & 50.29 & 40.36 & 56.21 & 49.89 & 48.25 & 53.76 & 38.64 & 45.78 & 46.61 \\
DR-Adapter~\cite{su2024DR-Adapter} & CVPR'24 & 53.66 & 60.06 & 67.01 & 70.28 & 62.75 & 54.43 & 62.52 & 54.15 & 40.81 & 52.98 \\
IFA$_{T=3}$~\cite{nie2024IFA} & CVPR'24 & 56.14 & 63.36 & 71.58 & 73.75 & 66.21 & 55.31 & 68.11 & 51.23 & \underline{46.04} & 55.17 \\
DIFD~\cite{cheng2025DIFD} & TMI'25 & \underline{57.41} & \underline{66.31} & \underline{75.17} & \underline{77.64} & \underline{69.13} & \underline{57.02} & \underline{74.08} & \underline{58.18} & 42.45 & \underline{57.93} \\

\midrule

\rowcolor{gray!20}
\textbf{Ours} & — & \textbf{76.08} & \textbf{72.27} & \textbf{84.86} & \textbf{88.95} & \textbf{80.54} & \textbf{77.82} & \textbf{82.97} & \textbf{71.22} & \textbf{68.07} & \textbf{74.82} \\
\bottomrule
\end{tabular}
% }
}
\end{table*}

\begin{table*}[htb]
    \caption{Comparison on the Abd-MRI under setting 2.}
    \label{tab:comparison2}
    \centering
    % \small
    
    % \setlength{\tabcolsep}{2pt}{
    % \resizebox{0.6\linewidth}{!}{
    
    \begin{tabular}{@{}l ccccc | ccc@{}}
        \toprule
        
        Method & Spleen & Liver & LK & RK & Mean$\uparrow$ & HD95$\downarrow$ & ASSD$\downarrow$ &  sample/s \\
        
        \midrule

        UniverSeg & 44.27 & 55.08 & 41.50 & 40.03 & 45.22 & 42.84 & 20.05 & \textbf{0.0080} \\ 
        MultiverSeg & 61.42 & 70.03 & 64.33 & 70.72 & 66.62 & 54.59 & 20.76 & 0.0728 \\ 
        DiffewS & 73.11 & 77.16 & 77.41 & 83.47 & 77.79 & 17.37 & 8.74 & 0.0768 \\ 
        \rowcolor{gray!20}
        \textbf{Ours} & \textbf{77.25} & \textbf{78.58} & \textbf{85.03} & \textbf{88.27} & \textbf{82.28} & \textbf{13.15} & \textbf{7.38} & 0.0914 \\
        
        % \midrule

        %  GT masks & 84.09 & 92.80 & 88.36 & 92.59 & 89.47 & 6.42 & 5.18 & 0.0923 \\
        
        \bottomrule
    \end{tabular}
    
    % }
    % }
    
\end{table*}

\subsection{Results under Setting 2 of CD-FSMIS}
We supplement the experiments on Setting 2 of Cross-Domain Few-Shot Medical Image Segmentation (CD-FSMIS), where the results of other comparative methods are directly adopted from DIFD~\cite{cheng2025DIFD}. As presented in \cref{tab:result_cross_domain_2}, the performance of our method under Setting 2 is comparable to that obtained under Setting 1 in the main text, and it outperforms the method proposed in DIFD by a significant margin across all cross-domain scenarios. Specifically, for the cross-modality task of Abd-CT → Abd-MRI, our method achieves a mean Dice score of 80.54\%, which represents a substantial improvement of 11.41\% over the 69.13\% achieved by DIFD. For the reverse cross-modality task of Abd-MRI → Abd-CT, our method yields an even larger performance gain of 16.89\% compared with DIFD. These results fully demonstrate that our method possesses more stable generalization capabilities and can better tackle the cross-domain challenges in few-shot medical image segmentation, maintaining high segmentation accuracy when transferring across different medical imaging modalities.

\begin{table*}[htb]
    \caption{Comparison on the Abd-CT under setting 2.}
    \label{tab:comparison}
    \centering
    % \small
    
    % \setlength{\tabcolsep}{2pt}{
    % \resizebox{0.6\linewidth}{!}{
    
    \begin{tabular}{@{}l ccccc | ccc@{}}
        \toprule
        
        Method & Spleen & Liver & LK & RK & Mean$\uparrow$ & HD95$\downarrow$ & ASSD$\downarrow$ &  sample/s \\
        
        \midrule

        UniverSeg & 34.58 & 51.22 & 31.07 &31.97 & 37.20 & 53.86 & 24.29 & \textbf{0.0077} \\ 
        MultiverSeg & 62.39 & 76.76 & 54.19 & 53.93 & 61.82 & 57.63 & 24.36 & 0.0551 \\ 
        DiffewS & 76.84 & 79.57 & 69.70 & 73.62 & 74.93 & 18.86 & 9.49 & 0.0732 \\ 
        \rowcolor{gray!20}
        \textbf{Ours} & \textbf{83.08} & \textbf{82.59} & \textbf{82.22} & \textbf{85.10} & \textbf{83.25} & \textbf{13.20} & \textbf{7.62} & 0.0856 \\

        % \midrule

        %  \textcolor{purple}{GT masks *} & 88.10 & 91.27 & 81.45 & 82.76 & 85.90 & 7.09 & 4.84 & 0.0828 \\ 
        
        \bottomrule
    \end{tabular}
    
    % }
    % }

\end{table*}

\begin{figure*}[htb]
  \centering
  \includegraphics[width=0.8\linewidth]{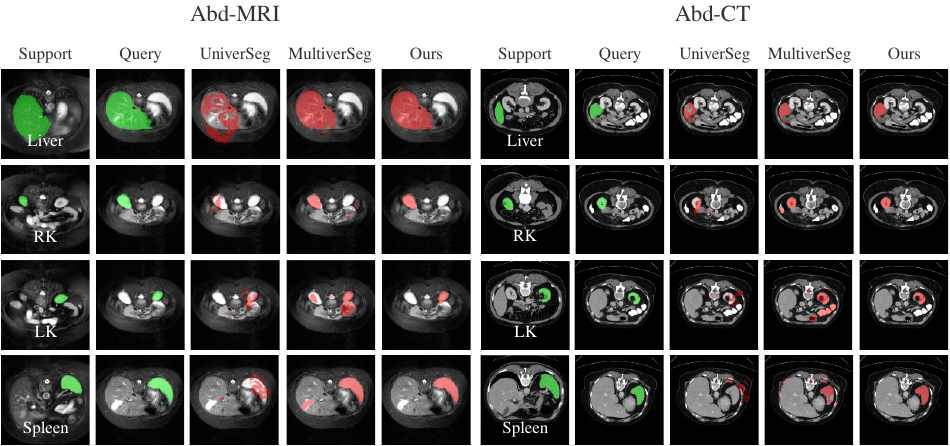}
  \caption{Qualitative comparison between our method and the universal models method on the Abd-MRI dataset and Abd-CT dataset.}
  \label{fig:vis_result_supp}
\end{figure*}

\subsection{Comparison with Universal Models}
Prior work did not compare against universal models or report HD95/ASSD. We therefore conducted additional experiments under Setting 2 with an input resolution of 256. 
% We also included an extra reference baseline where our method is supervised with GT masks.

Shown in \cref{tab:comparison2} and \cref{tab:comparison}, SD-FSMIS significantly outperforms universal models. On Abd-CT, it exceeds UniverSeg~\cite{butoi2023universeg} and MultiverSeg~\cite{wong2025multiverseg} by +46.05\% and +21.43\% in mean Dice, respectively. On Abd-MRI, the gains are +37.06\% and +15.66\%. Universal models often fail to distinguish visually similar background tissues, leading to confused masks and high HD95, whereas our method produces more accurate boundaries. 

\textbf{Efficiency.}
Our method adopts single-step denoising, directly generating the segmentation from timestep $t = 999$, which reduces inference cost. The resulting inference time is 0.09s per image, remaining within the real-time range, although slower than UniverSeg and MultiverSeg. Importantly, this minor latency increase brings substantial gains in accuracy and cross-domain robustness, which is the core contribution of our work. In medical imaging, robustness and precision are significantly more critical than minimal latency. We therefore consider this trade-off both practical and clinically reasonable.

\textbf{Visualization.}
As illustrated in \cref{fig:vis_result_supp}, we conduct 1-shot segmentation experiments on Abd-MRI and Abd-CT, and compare our method with two representative universal segmentation models: UniverSeg and MultiverSeg. For UniverSeg, under the 1-shot support set setting, the model almost fails to perform effective segmentation of the target organs. Its generated masks are randomly scattered around the target regions with no clear structural consistency, which reflects the poor adaptability of vanilla universal models to medical image segmentation tasks with limited annotated samples. MultiverSeg shows an improved performance compared to UniverSeg and can roughly localize the target organs in both Abd-MRI and Abd-CT images. However, this model still suffers from obvious limitations in fine-grained boundary segmentation: it cannot accurately distinguish the foreground target organs from the visually similar background tissues (e.g., adjacent visceral tissues and parenchyma), thus resulting in frequent under-segmentation (missing partial valid regions of target organs) and over-segmentation (erroneously including background tissues into the segmentation masks) issues.

In contrast, our method achieves more robust and accurate segmentation in the 1-shot scenario. It not only precisely localizes the target organs but also effectively discriminates the subtle boundary differences between the foreground organs and the background tissues with similar visual features. 
This superior performance is attributed to our method leveraging the pre-trained Stable Diffusion model with strong visual priors, which we adapt to the few-shot medical image segmentation task via dedicated design. This adaptation unlocks the model’s powerful generalization capability and equips it with a stronger ability to capture organ-specific anatomical features and discriminate visually similar tissues.

\begin{figure}[htb]
  \centering
  \includegraphics[width=\linewidth]{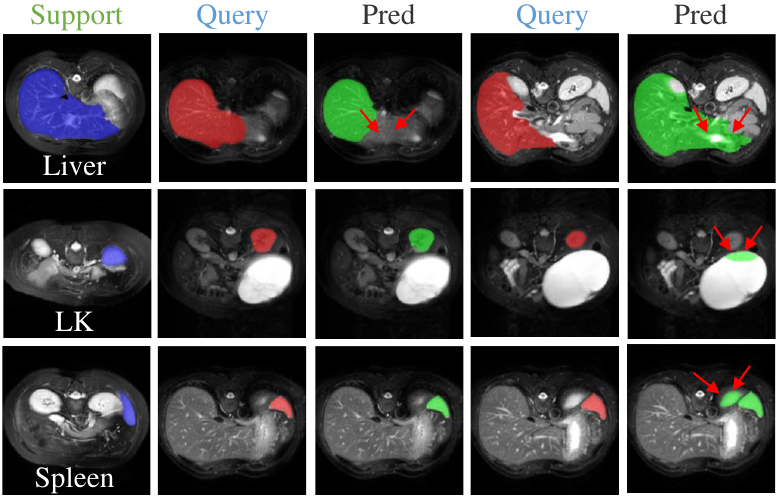}
  \caption{Visualization of failure cases on the Abd-MRI dataset.}
  \label{fig:analyse}
\end{figure}

% \begin{figure}[t]
%   \centering
%   \fbox{\rule{0pt}{2in} \rule{0.9\linewidth}{0pt}}
%    %\includegraphics[width=0.8\linewidth]{egfigure.eps}

%    \caption{Example of caption.
%    It is set in Roman so that mathematics (always set in Roman: $B \sin A = A \sin B$) may be included without an ugly clash.}
%    \label{fig:onecol}
% \end{figure}

\section{Analysis and Visualization}
% 分析
\subsection{Analysis Failure Cases}
\label{sub:analysis}
Despite the overall effectiveness of SD-FSMIS, our evaluation on the Abd-MRI dataset reveals some performance discrepancies, particularly for certain organ classes. As illustrated in \cref{fig:analyse}, visual inspection of the segmentation results indicates that the model occasionally produces incomplete or over-segmented masks for the Liver. This issue appears to stem from the inherently low contrast between the liver tissue and the surrounding background, resulting in ambiguous boundaries that challenge the model’s ability to distinguish between foreground and background regions.

In addition, when segmenting the Left Kidney (LK), we observe that the model's attention may be disproportionately drawn to regions with extreme saliency in a given image slice. This can lead to inconsistent performance across consecutive slices—while one slice may be segmented accurately, the subsequent slice might exhibit segmentation errors, misidentifying the target region. 

Furthermore, in cases where both the Spleen and the LK appear within the same slice and are positioned in close proximity, the model tends to merge these adjacent organs into a single segmentation output. These mis-segmentation events suggest that the spatial relationships and relative proximities between organs play a critical role in challenging the model's discriminative capacity.

These findings highlight specific challenges in medical image segmentation, where subtle contrast differences and complex anatomical interactions can lead to segmentation inaccuracies.
Addressing these issues by enhancing attention mechanisms or improving boundary detection strategies may further improve the robustness of SD-FSMIS in future work.
% Addressing these issues—potentially through enhanced attention mechanisms or refined boundary detection strategies—may further improve the robustness of SD-FSMIS in future work.

% \input{fig/analyse}
% \input{fig/denoise}

\subsection{Visualization of Training}
\begin{figure}[!ht]
  \centering
  \includegraphics[width=\linewidth]{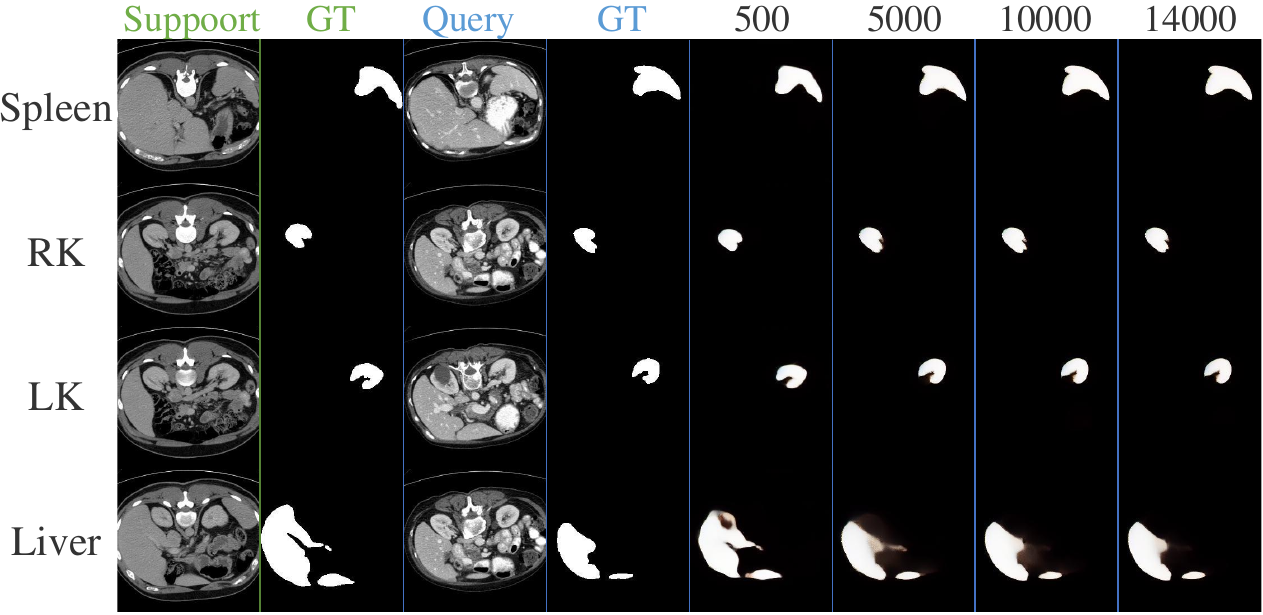}
  \caption{Visualization of the training process.}
  \label{fig:denoise}
\end{figure}

Additionally, \cref{fig:denoise} illustrates the performance of our method during the training process on the Abd-CT dataset. Notably, even in the early stages of training (after 500 iterations), the model is able to segment simpler classes effectively, and for more complex organs such as the liver, good segmentation results are achieved as early as 5,000 iterations. These findings further underscore the powerful capabilities of diffusion models in tackling few-shot segmentation challenges.

% ----------- Supplementary Content Ends Here -----------

% References and End of Paper
% These lines must be placed at the end of your paper
% {
%     \small
%     \bibliographystyle{ieeenat_fullname}
%     \bibliography{ref}
% }

{
    \small
    \bibliographystyle{ieeenat_fullname}
    \bibliography{ref}
}

\end{document}